\documentclass{article}
\usepackage{spconf,amsmath,graphicx}
\usepackage{bm} 
\usepackage{bbding}
\usepackage{array}
\usepackage{amssymb}
\usepackage{multirow}
\usepackage{booktabs}

\title{MSSTNet: A Multi-Scale Spatio-Temporal CNN-Transformer Network for Dynamic Facial Expression Recognition}
\name{Linhuang Wang$^{\star}$, Xin Kang$^{\star}$, Fei Ding$^{\star}$\qquad Satoshi Nakagawa$^{\dagger}$\qquad Fuji Ren$^{\star \dagger}$ \thanks{This research has been supported by the Project of Discretionary Budget of the Dean, Graduate School of Technology, Industrial and Social Sciences, Tokushima University.}}

\address{$^{\star}$ Advanced Technology and Science, University of Tokushima, Tokushima, Japan \\
	$^{\dagger}$Graduate School of Information Science \& Technology, The University of Tokyo, Tokyo, Japan \\
	$^{\star \dagger}$ School of Computer Science and Engineering, \\ University of Electronic Science and Technology of China, Chengdu, China}
\begin{document}
	%
	\maketitle
	\begin{abstract}
		Unlike typical video action recognition, Dynamic Facial Expression Recognition (DFER) does not involve distinct moving targets but relies on localized changes in facial muscles. Addressing this distinctive attribute, we propose a Multi-Scale Spatio-temporal CNN-Transformer network (MSSTNet). Our approach takes spatial features of different scales extracted by CNN and feeds them into a Multi-scale Embedding Layer (MELayer). The MELayer extracts multi-scale spatial information and encodes these features before sending them into a Temporal Transformer (T-Former). The T-Former simultaneously extracts temporal information while continually integrating multi-scale spatial information. This process culminates in the generation of multi-scale spatio-temporal features that are utilized for the final classification. Our method achieves state-of-the-art results on two in-the-wild datasets. Furthermore, a series of ablation experiments and visualizations provide further validation of our approach's proficiency in leveraging spatio-temporal information within DFER.
	\end{abstract}
	\section{Introduction}
	\label{sec:intro}
	
	Facial expression recognition (FER) finds extensive applications in daily life, such as affective computing, human-computer interaction, fatigue detection in driving, and emotional robots, among others \cite{wang2023centermatch}. Dynamic Facial Expression Recognition (DFER) aims to recognize emotions within a video segment, which typically yields more precise results than single static images.
	
	Traditional DFER methods rely on manual feature extraction techniques, such as LBP-TOP \cite{dhall2013emotion}, STLMBP \cite{huang2014improved}, and HOG-TOP \cite{chen2014emotion}. With the rise of deep learning, models based on deep learning have emerged as the mainstream approach in the DFER domain. These approaches utilize the Convolutional Neural Network (CNN) to extract image features, followed by modeling of the temporal sequences through models like Recurrent Neural Network (RNN)~\cite{wang2019multi,ebrahimi2015recurrent}, Long Short-Term Memory (LSTM)~\cite{hochreiter1997long}, or Gated Recurrent Unit (GRU) \cite{chung2014empirical}. Additionally, 3D CNNs ~\cite{fan2016video} are commonly employed for DFER as they can simultaneously capture spatial and temporal features. In recent years, the Transformer \cite{vaswani2017attention} architecture, which has achieved significant success in the Natural Language Processing (NLP) field, has gradually been introduced to computer vision. Its global attention mechanism and parallel computing capabilities are effective for modeling time sequences. Consequently, an increasing number of researchers are incorporating transformers into DFER~\cite{zhao2021former,li2022nr}, and some even explore their use in methods designed for video action recognition~\cite{bertasius2021space,li2021self,wang2021anchor}. 
	
	\begin{figure}[t]
		\centering
	\includegraphics[width=0.8\linewidth]{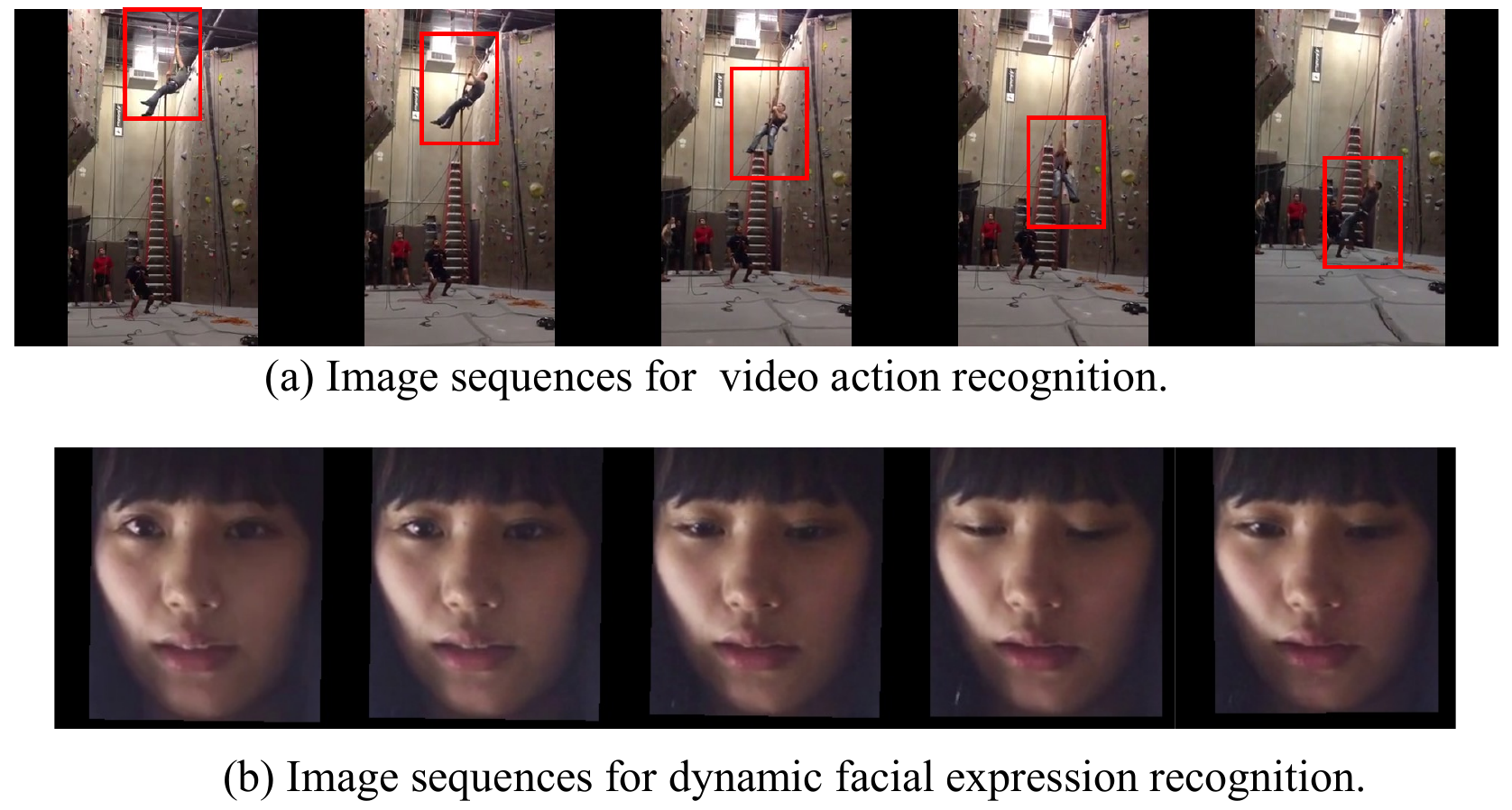}
		\caption{(a) and (b) represent image sequences for video action recognition and DFER, respectively. In (a), distinct moving targets are present, while in (b), only localized changes in facial muscle states are observed.}
		\label{fig:tu1}
	\end{figure}
	
	\begin{figure*}[t]
		\begin{center}
			\includegraphics[width=0.95\textwidth]{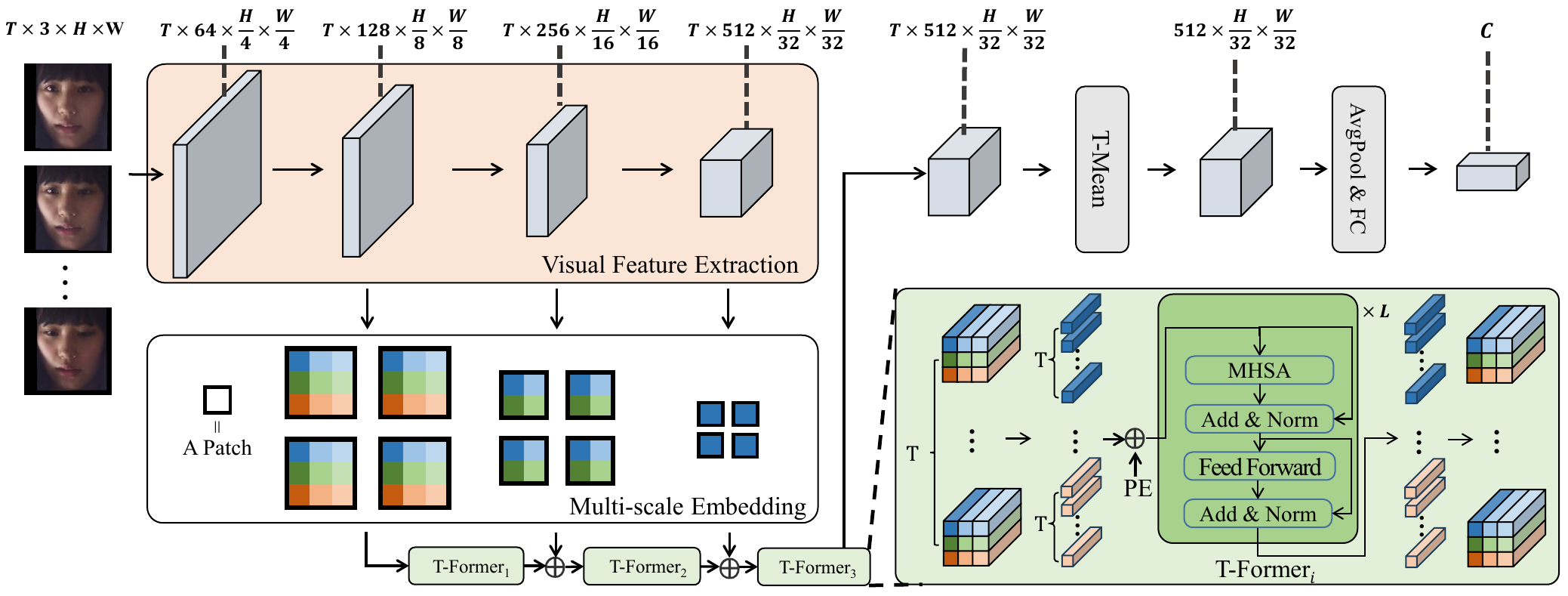}
		\end{center}
		\caption{The overall architecture of MSSTNet. The visual feature extraction is a CNN backbone network, and the T-Former is a transformer structure composed of $L$ blocks. PE signifies temporal positional embedding, $\bigoplus$ denotes element-wise addition, MHSA stands for multi-head self-attention mechanism and T-Mean indicates temporal dimension averaging.} 
		\label{zongtu}
	\end{figure*}
	
	However, unlike typical video action recognition, DFER does not involve clearly moving targets. As depicted in Figure \ref{fig:tu1}(a), video action recognition often  deals with prominently shifting targets. In contrast,  in the context of DFER, as illustrated in Figure \ref{fig:tu1}(b), the image sequences typically undergo facial alignment preprocessing. Consequently, there are no moving targets; instead, the variations observed are in the state of facial muscles over time \cite{ma2023logo}.
	
	In response to these characteristics, we propose the Multi-Scale Spatio-temporal CNN-Transformer Network (MSSTNet). We introduce a novel Multi-scale Embedding Layer (MELayer), which encodes spatial features of varying sizes extracted by the CNN, and subsequently feeds them into the Temporal Transformer (T-Former). The T-Former, while extracting temporal information, simultaneously integrates spatial information of different scales, culminating in the generation of multi-scale spatio-temporal features. Our method achieves state-of-the-art performance on two widely-used in-the-wild DFER datasets. A series of ablation experiments and visualizations provide empirical evidence of the effectiveness of the spatio-temporal features generated by our approach.
	
	\section{The Proposed Method}
	The overall architecture of our approach is depicted in Figure \ref{zongtu}. Given an image sequence $\bm{X} \in \mathbb{R}^{T\times 3\times H\times W}$ with $T$ RGB facial frames of size $H\times W$, we employ a CNN backbone to extract visual features of varying scales, including spatial positional information. The MELayer encodes these features and passes them to the T-Former. The T-Former utilizes the encoder architecture from the Transformer to extract temporal information while concurrently integrating multi-scale spatial information. The T-Former's output is subjected to temporal and spatial dimension-wise averaging, followed by a fully connected network, to obtain the final classification result.

	\subsection{Multi-scale Embedding Layer (MELayer)}
	The MELayer takes feature maps of varying sizes and encodes each of them into $N$ patches. A particular input denoted as $\bm{F} \in \mathbb{R}^{T\times d\times H'\times W'}$, comprising $T$ feature maps, with each feature map having $d$ channels and spatial dimensions $H'\times W'$. We decompose each feature map into $N$ non-overlapping patches, each of size $P \times P$, such that the $N$ patches span the entire frame, i.e., $\smash{P^2=H'W'/N}$. We flatten these patches into vectors $\smash{{\bf x}_{(p,t)} \in \mathbb{R}^{P^2 d}}$ with $p=1,\hdots,N$ denoting spatial locations and $t=1,\hdots,T$ depicting an index over frames. We linearly map each patch $\smash{{\bf x}_{(p,t)}}$ into an embedding vector $\smash{{\bf z}_{(p,t)}^{(0)} \in \mathbb{R}^{D}}$ by means of a learnable matrix $\smash{E \in \mathbb{R}^{D  \times P^2 d}}$:
	\vspace{-0.3cm}
	\begin{equation}
		{\bf z}_{(p,t)}^{(0)} = E {\bf x}_{(p,t)} + {\bf e}_{t}^{pos},
	\end{equation}
	where $\smash{{\bf e}_{t}^{pos} \in \mathbb{R}^{D}}$ represents a learnable temporal positional embedding. It's worth noting that spatial positional embeddings are not included here. Since the T-Former treats the same region as tokens, spatial positional embedding is unnecessary. Consequently, in this context, only temporal position embeddings are introduced. The resulting sequence of embedding vectors $\smash{{\bf z}_{(p,t)}^{(0)}}$ represents the input to the T-Former. 
	
	\subsection{Temporal Transformer (T-Former)}
	The T-Former consists of $L$ Transformer encoding blocks, designed to extract temporal features of the same region at different time points while progressively integrating multi-scale spatial information.
	
	\textbf{Query-Key-Value Computation.} At each block $\ell$, a query/key/value vector is computed for each patch from the representation $\smash{{\bf z}_{(p,t)}^{(\ell-1)}}$ encoded by the preceding block:
	\vspace{-0.3cm}
	\begin{align}
		{\bf q}_{(p,t)}^{(\ell,a)} &= W_Q^{(\ell,a)} \mathrm{LN}\left({\bf z}_{(p,t)}^{(\ell-1)}\right) \in \mathbb{R}^{D_h},\\
		{\bf k}_{(p,t)}^{(\ell,a)} &= W_K^{(\ell,a)} \mathrm{LN}\left({\bf z}_{(p,t)}^{(\ell-1)}\right) \in \mathbb{R}^{D_h},\\
		{\bf v}_{(p,t)}^{(\ell,a)} &= W_V^{(\ell,a)} \mathrm{LN}\left({\bf z}_{(p,t)}^{(\ell-1)}\right) \in \mathbb{R}^{D_h},
	\end{align}
     where $\mathrm{LN}()$ denotes LayerNorm, $a = 1, \hdots, \mathcal{A}$ is an index over multiple attention heads and $\mathcal{A}$ denotes the total number of attention heads. The latent dimensionality for each attention head is set to $D_h = D / \mathcal{A}$.
	
	\textbf{Self-attention Computation.} Self-attention weights are computed via dot-product. The self-attention weights $\smash{\pmb{\alpha}_{(p,t)}^{(\ell,a)} \in \mathbb{R}^{T}}$ for query patch $(p,t)$ are given by:
	\vspace{-0.3cm}
	\begin{align}
		\pmb{\alpha}_{(p,t)}^{(\ell,a)} &= \mathrm{SM}\left( \frac{{\bf q}_{(p,t)}^{(\ell,a)}}{\sqrt{D_h}}^\top \cdot  \left\{ {\bf k}_{(p,t')}^{(\ell,a)} \right\}_{t'=1,...,T}   \right),
	\end{align}
	where $\mathrm{SM}$ denotes the softmax activation function. It's worth noting that in our approach, we exclusively compute self-attention in the temporal dimension. This choice results in a significant reduction in computational cost compared to simultaneously computing attention in both spatial and temporal dimensions.
	
	\textbf{Encoding.} The encoding $\smash{{\bf z}_{(p,t)}^{(\ell)}}$ at block $\ell$ is obtained by first computing the weighted sum of value vectors using self-attention coefficients from each attention head:
	\vspace{-0.3cm}
	\begin{align}
		{\bf s}_{(p,t)}^{(\ell,a)} &=  \sum_{t'=1}^T {\alpha}_{(p,t),(p,t')}^{(\ell,a)} {\bf v}_{(p,t')}^{(\ell,a)}.
	\end{align}
	\vspace{-0.5cm}
	
	Then, the concatenation of these vectors from all heads is projected and passed through an MLP, using residual connections after each operation:
	
	\vspace{-0.3cm}
	\begin{align}
		{\bf z'}_{(p,t)}^{(\ell)} &= W_O \left[ \begin{array}{c} {\bf s}_{(p,t)}^{(\ell,1)}\\ \vdots \\ {\bf s}_{(p,t)}^{(\ell,\mathcal{A})} \end{array} \right] + {\bf z}_{(p,t)}^{(\ell-1)} \label{eq:hiddenstate},\\
		{\bf z}_{(p,t)}^{(\ell)} &= \mathrm{MLP}\left(\mathrm{LN}\left({\bf z'}_{(p,t)}^{(\ell)}\right)\right) + {\bf z'}_{(p,t)}^{(\ell)}.
	\end{align}

	All the features produced by the final block, denoted as $\smash{{\bf z}_{(p,t)}^{(L)}}$ are averaged across both the temporal and spatial dimensions to obtain $\bar{z}\in \mathbb{R}^{D}$, followed by further processing through a fully connected (FC) layer to obtain the final result.
	\vspace{-0.3cm}
	\begin{equation}
		\bm{p} = FC(\bar{z}),
	\end{equation}
	where $\bm{p} \in \mathbb{R}^{C}$ is the prediction distribution of facial expression classes.
	
	\begin{table}[!t]
		\caption{Evaluation of each component and different settings for MSSTNet. $T$ represents the number of input frames, and the numbers in the T-Former$_i$ column denote the quantity of blocks. The best results for each frame number are in bold.}
		\centering
		\resizebox{0.8\linewidth}{!}{
			\begin{tabular}{c|ccc|cc}
				\toprule
				\multirow{2}{*}{$T$} & \multirow{2}{*}{T-Former$_1$} &  \multirow{2}{*}{T-Former$_2$} & \multirow{2}{*}{T-Former$_3$} & \multicolumn{2}{c}{WAR(\%)} \\
				\cmidrule{5-6}
				~ & ~ & ~ & ~ & FERV39k & DFEW        \\
				\midrule
				\multirow{7}{*}{4} & 
				~ & ~ & ~ & 46.89 & 65.91 \\
				~ & ~ & ~ & $\checkmark$ & 47.83 & 66.68 \\
				~ & $\checkmark$ & ~ & $\checkmark$  & 48.08 & 67.93 \\
				~ & ~ & $\checkmark$ & $\checkmark$  & 48.42 & 68.37 \\
				\cmidrule{2-6}
				~ & 1 & 1 & 1 & 49.94 & 68.97 \\
				~ & 2 & 2 & 2 & \textbf{49.89} & \textbf{69.71} \\
				~ & 3 & 3 & 3 & 49.11 &	69.36 \\
				\midrule
				\midrule
				\multirow{3}{*}{8} 
				~ & 1 & 1 & 1 & 49.66 & 69.89\\
				~ & 2 & 2 & 2 & \textbf{50.25} & \textbf{70.16} \\
				~ & 3 & 3 & 3 & 49.29 & 69.65 \\
				\midrule
				\midrule
				\multirow{3}{*}{12} 
				~ & 1 & 1 & 1 & 49.97 & 70.53 \\
				~ & 2 & 2 & 2 & \textbf{50.43} & \textbf{70.85} \\
				~ & 3 & 3 & 3 & 50.01 & 70.31 \\		
				\midrule
				\midrule
				\multirow{3}{*}{16} 
				~ & 1 & 1 & 1 & 50.17 & 70.81 \\
				~ & 2 & 2 & 2 & \textbf{51.02} & \textbf{71.42} \\
				~ & 3 & 3 & 3 & 50.42 & 70.99 \\
				\bottomrule            
			\end{tabular}
		}
		\label{table:components_study}
	\end{table}

	\begin{table}[!t]
		\caption{Comparison with state-of-the-art methods on FERV39k. The best results are in bold.}
		\centering
		\resizebox{0.7\linewidth}{!}{
			\begin{tabular}{c|cc}
				\toprule
				\multirow{2}{*}{Method} & \multicolumn{2}{c}{Metrics(\%)}  \\ \cmidrule{2-3} 
				& WAR   & UAR   \\ 
				\midrule
				R18-LSTM\cite{he2016residual,hochreiter1997long,wang2022ferv39k}  & 42.59 & 30.92 \\
				VGG13-LSTM\cite{simonyan2014very,hochreiter1997long,wang2022ferv39k} & 43.37 & 32.41 \\
				C3D\cite{tran2015learning,wang2022ferv39k} & 31.69 & 22.68 \\
				3D ResNet18\cite{tran2018closer,wang2022ferv39k} & 37.57 & 26.67 \\
				Former-DFER\cite{zhao2021former}             & 45.72 & 36.88   \\ 
				LOGO-Former\cite{ma2023logo} & 48.13 & 38.22 \\ 
				\midrule
				MSSTNet$_{T=4}$(Ours) & 49.89 & 39.80 \\
				MSSTNet$_{T=8}$(Ours) & 50.25 & 40.46 \\
				MSSTNet$_{T=12}$(Ours) & 50.43 & 40.70 \\
				MSSTNet$_{T=16}$(Ours)     & \multicolumn{1}{c}{\textbf{51.02}} & \multicolumn{1}{c}{\textbf{41.28}} \\ 
				\bottomrule
			\end{tabular}
		}
		\label{table:FERV_result}
	\end{table}
	
	\begin{table*}[!t]
			\caption{Comparison with state-of-the-art methods on DFEW. \textbf{Bold} denotes the best result. \underline{Underline} denotes the second best result.}
			\centering
			\resizebox{0.85\textwidth}{!}{
				\begin{tabular}{c|ccccccc|cc|c}
					\toprule
					\multirow{2}{*}{Method}&
					\multicolumn{7}{c|}{Accuracy of Each Emotion (\%) }&\multicolumn{2}{c|}{ Metrics (\%)}&\multirow{2}{*}{FLOPs} \\
					\cmidrule(lr){2-10}&
					Happy & Sad & Neutral & Angry & Surprise & Disgust & Fear & UAR & WAR&(G) \\
					\midrule
					R(2+1)D18 \cite{tran2018closer} &79.67 &39.07 &57.66 &50.39 &48.26 &3.45 &21.06 &42.79 &53.22&42.36 \\
					3D Resnet18 \cite{hara2018can}&73.13 &48.26 &50.51 &64.75 &50.10 &0.00 &26.39 &44.73 &54.98&8.32 \\
					ResNet18+LSTM \cite{hochreiter1997long,he2016residual}&78.00 &40.65 &53.77 &56.83 &45.00 &4.14 &21.62 &42.86 &53.08&7.78 \\
					Resnet18+GRU \cite{he2016residual,chung2014empirical}&82.87 &63.83 &65.06 &68.51 &52.00 &0.86 &30.14 &51.68 &64.02&7.78 \\
					Former-DFER \cite{zhao2021former}&84.05 &62.57 &67.52 &70.03 &56.43 &3.45 &31.78 &53.69 &65.70&9.11 \\
					DPCNet \cite{wang2022dpcnet}&- &- &- &- &- &- &- &57.11 &66.32&9.52 \\
					NR-DFERNet \cite{li2022nr}& 86.42 & 65.10 & 70.40 &72.88&50.10&0.00&\textbf{45.44}&55.77&68.01&\underline{6.33} \\
					LOGO-Former \cite{ma2023logo}& 85.39 & 66.52 & 68.94 & 71.33 & 54.59 & 0.00 & 32.71 & 54.21 & 66.98 & 10.27 \\
					\midrule
					MSSTNet$_{T=4}$(Ours) &89.57 &\underline{71.28} &69.01 &70.92 &55.89 &11.03 &43.68 &58.77 &69.71&\textbf{5.05} \\
					MSSTNet$_{T=8}$(Ours) &89.20 &70.12 &\textbf{72.35} &70.23 &59.22 &\underline{13.10} &39.02 &59.04 &70.16&10.10 \\
					MSSTNet$_{T=12}$(Ours) &\underline{90.22} &70.22 &70.93 &\textbf{74.60}&\underline{59.50} &10.34 &38.58 &\underline{59.20} &\underline{70.85}&15.15 \\
					MSSTNet$_{T=16}$(Ours)&\textbf{90.39}&\textbf{71.54}&\underline{71.94}&\underline{73.17}&\textbf{61.20}&\textbf{13.79}&\underline{39.91}&\textbf{60.28}&\textbf{71.42}&20.20\\
					\bottomrule
				\end{tabular}
			}
		\label{table:DFER_result}
	\end{table*}

	\section{Experiments}
	We conducted experiments on two widely-used real-world datasets: \textbf{DFEW} \cite{jiang2020dfew} and \textbf{FERV39k} \cite{wang2022ferv39k}, both of which provide images that have undergone official facial alignment processing. We employed the Weighted Average Recall (WAR) and the Unweighted Average Recall (UAR) as evaluation metrics. Following prior research \cite{zhao2021former,li2022nr,ma2023logo}, we utilized ResNet-18 as the CNN backbone. In the T-Former architecture, we configured the feature dimension $D$, number of patches $N$ and number of heads as 768, 16 and 8, respectively. All our models were implemented using PyTorch and trained on an NVIDIA TITAN RTX GPU. The training process employed the SGD optimizer with an initial learning rate of 0.1 for 40 epochs. Learning rate decay by a factor of 10 occurred at the 20th and 35th epochs.
	
	\begin{figure}[!tb]
		\begin{center}
			\includegraphics[width=0.75\linewidth]{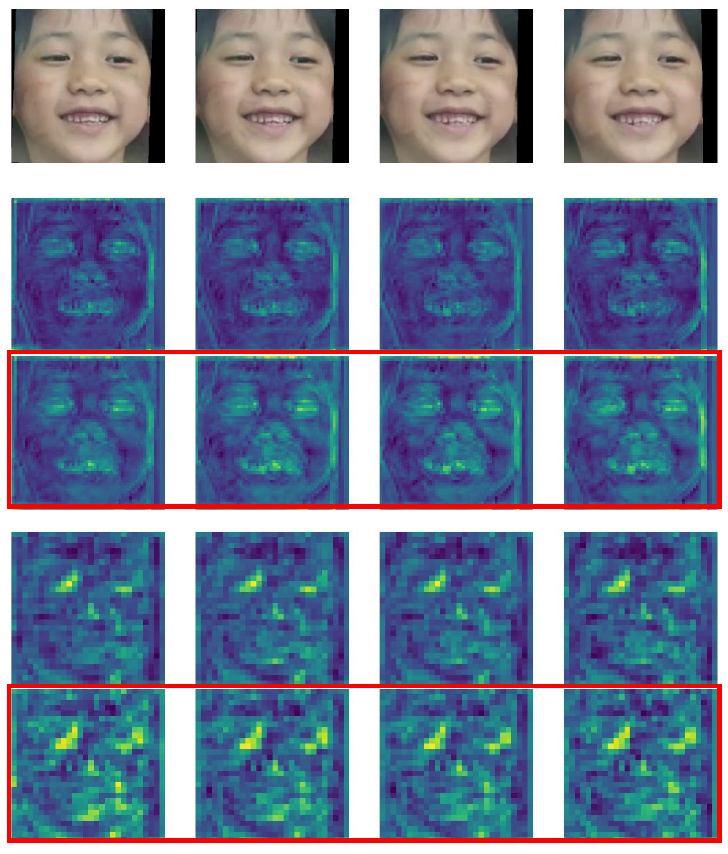}
		\end{center}
		\caption{Visualization of feature map. The feature maps within the red boxes represent the outputs after passing through the T-Former, while those without boxes depict feature maps without undergoing the T-Former.}
		\label{fig:feature}
	\end{figure}
	
	\textbf{Ablation Studies.} We evaluated the impact of various components of MSSTNet, and the experimental results are presented in Table \ref{table:components_study}. Each stage of the T-Former contributes to an enhancement in model performance, affirming the T-Former's proficiency in capturing temporal features. Notably, the combined interaction of all T-Former stages yields the greatest improvement in model performance, underscoring the effectiveness of multi-scale spatio-temporal features. Moreover, merely increasing the number of T-Former blocks does not lead to improved performance; instead, it introduces additional computational complexity. As a result, we set the number of blocks to 2 in subsequent experiments. Furthermore, the model's performance improves with an increase in the number of input frames, validating the efficacy of our approach in modeling the temporal dimension.

	\textbf{Comparison with State-of-the-Art Methods.} The comparative analysis between our method and others on the two datasets is presented in Tables \ref{table:FERV_result} and \ref{table:DFER_result}. Our approach outperforms previous methods on both datasets, regardless of the number of input frames. On the \textbf{DFEW} dataset, MSSTNet outperforms the previous best method with significant improvements of 3.41\% in WAR and 3.17\% in UAR, with a notable 9.65\% increase in accuracy for the \textit{disgust} category. Furthermore, MSSTNet marks a pioneering achievement as the first method to attain a WAR of over 70\%. On the \textbf{FERV39k} dataset, our method also establishes a new benchmark by achieving a WAR exceeding 50\%. It surpasses the previous best method with improvements of 2.89\% in WAR and 3.06\% in UAR.

	\textbf{Visualization.} To evaluate the effectiveness of the T-Former in our approach, we conducted feature map visualizations as shown in Figure \ref{fig:feature}. The visualized outcomes reveal that, compared to feature maps without the T-Former, those processed by the T-Former demonstrate a clear focus on regions that reflect facial muscles variations, such as the eyes, nose, and mouth. This observation underscores the robust global temporal feature extraction capability of the T-Former, contributing to the model's enhanced performance.
	
	\section{Conclusion}
	In this paper, we address the dynamic variations within localized facial muscle regions in DFER by proposing a Multi-Scale Spatio-temporal CNN-Transformer network (MSSTNet). Our approach leverages CNN to capture spatial information, while Multi-scale Embedding Layer (MELayer) focuses on features within localized regions experiencing variations. Furthermore, a Temporal Transformer (T-Former) is employed to extract temporal information and integrate it with multi-scale spatial information, ultimately yielding multi-scale spatio-temporal features. Importantly, our method achieves state-of-the-art results on two commonly used in-the-wild datasets. Ablation experiments and visualizations further substantiate the effectiveness of our approach.

%


	\small
	\bibliographystyle{IEEEbib}
	\bibliography{refs}
	
\end{document}